\documentclass{article}
\PassOptionsToPackage{numbers}{natbib}

\usepackage[dblblindworkshop, final, nonatbib]{neurips_2025}
\usepackage{graphicx}
\usepackage{wrapfig}
\usepackage{xcolor}       % for coloring
\usepackage{amssymb}      % for symbols like \checkmark
\usepackage{pifont}       % for \ding symbols (\ding{55} for xmark)
\newcommand{\xmark}{\ding{55}}  % ✗
\usepackage{amsmath}
\workshoptitle{Machine Learning for the Physical Sciences}

\usepackage[utf8]{inputenc} % allow utf-8 input
\usepackage[T1]{fontenc}    % use 8-bit T1 fonts
\usepackage{hyperref}       % hyperlinks
\usepackage{url}            % simple URL typesetting
\usepackage{booktabs}       % professional-quality tables
\usepackage{amsfonts}       % blackboard math symbols
\usepackage{nicefrac}       % compact symbols for 1/2, etc.
\usepackage{microtype}      % microtypography
\usepackage{xcolor}         % colors

\title{Forecasting the Ionosphere from Sparse GNSS Data with Temporal-Fusion Transformers}

\author{%
  Giacomo Acciarini \\
  Advanced Concepts Team\\
  European Space Agency (ESA) \\
  \texttt{giacomo.acciarini@esa.int} \\
  \And
  Simone Mestici \\
  Department of Physics\\
  Università degli Studi di Roma Sapienza \\
  \texttt{simone.mestici@uniroma1.it} \\
  \And
  Halil S.~Kelebek \\
  Department of Engineering Science \\
  University of Oxford \\
  \texttt{halil@robots.ox.ac.uk} \\
  \And
  Linnea M.~Wolniewicz \\
  Department of Information and Computer Science \\
  University of Hawai'i at Mānoa\\
  \texttt{linneamw@hawaii.edu} \\
  \And
  Michael D.~Vergalla \\
  Free Flight Research Lab \\
  \texttt{mike@freeflightlab.org} \\
  \And
  Madhulika Guhathakurta \\
  NASA Headquarters \\
  \texttt{madhulika.guhathakurta@nasa.gov}
  \And  
  Umaa Rebbapragada \\
  NASA Jet Propulsion Laboratory \\
  \texttt{umaa.d.rebbapragada@jpl.nasa.gov}
  \And
  \And
  Bala Poduval \\
  University of New Hampshire \\
  \texttt{balapoduval@gmail.com }\\
  \And
  Atılım Güneş Baydin \\
  Department of Computer Science \\
  University of Oxford, UK \\
  \texttt{gunes@robots.ox.ac.uk} \\
  \And
  Frank Soboczenski \\
  Department of Computer Science \\
  University of York \& King’s College London\\
  \texttt{frank.soboczenski@york.ac.uk} \\
}

\begin{document}

\maketitle

\begin{abstract}

The ionosphere critically influences Global Navigation Satellite Systems (GNSS), satellite communications, and Low Earth Orbit (LEO) operations, yet accurate prediction of its variability remains challenging due to nonlinear couplings between solar, geomagnetic, and thermospheric drivers. Total Electron Content (TEC), a key ionospheric parameter, is derived from GNSS observations, but its reliable forecasting is limited by the sparse nature of global measurements and the limited accuracy of empirical models, especially during strong space weather conditions. In this work, we present a machine learning framework for ionospheric TEC forecasting that leverages Temporal Fusion Transformers (TFT) to predict sparse ionosphere data. Our approach accommodates heterogeneous input sources, including solar irradiance, geomagnetic indices, and GNSS-derived vertical TEC, and applies preprocessing and temporal alignment strategies. Experiments spanning 2010–2025 demonstrate that the model achieves robust predictions up to 24 hours ahead, with root mean square errors as low as 3.33 TECU. Results highlight that solar EUV irradiance provides the strongest predictive signals. Beyond forecasting accuracy, the framework offers interpretability through attention-based analysis, supporting both operational applications and scientific discovery. To encourage reproducibility and community-driven development, we release the full implementation as the open-source toolkit \texttt{ionopy}.
\end{abstract}

\section{Introduction}
\label{sec:introduction}
\begin{wrapfigure}{r}{0.3\linewidth} % right side, 35% width
    \centering
    \vspace{-10pt} % optional: remove vertical gap
    \includegraphics[width=\linewidth]{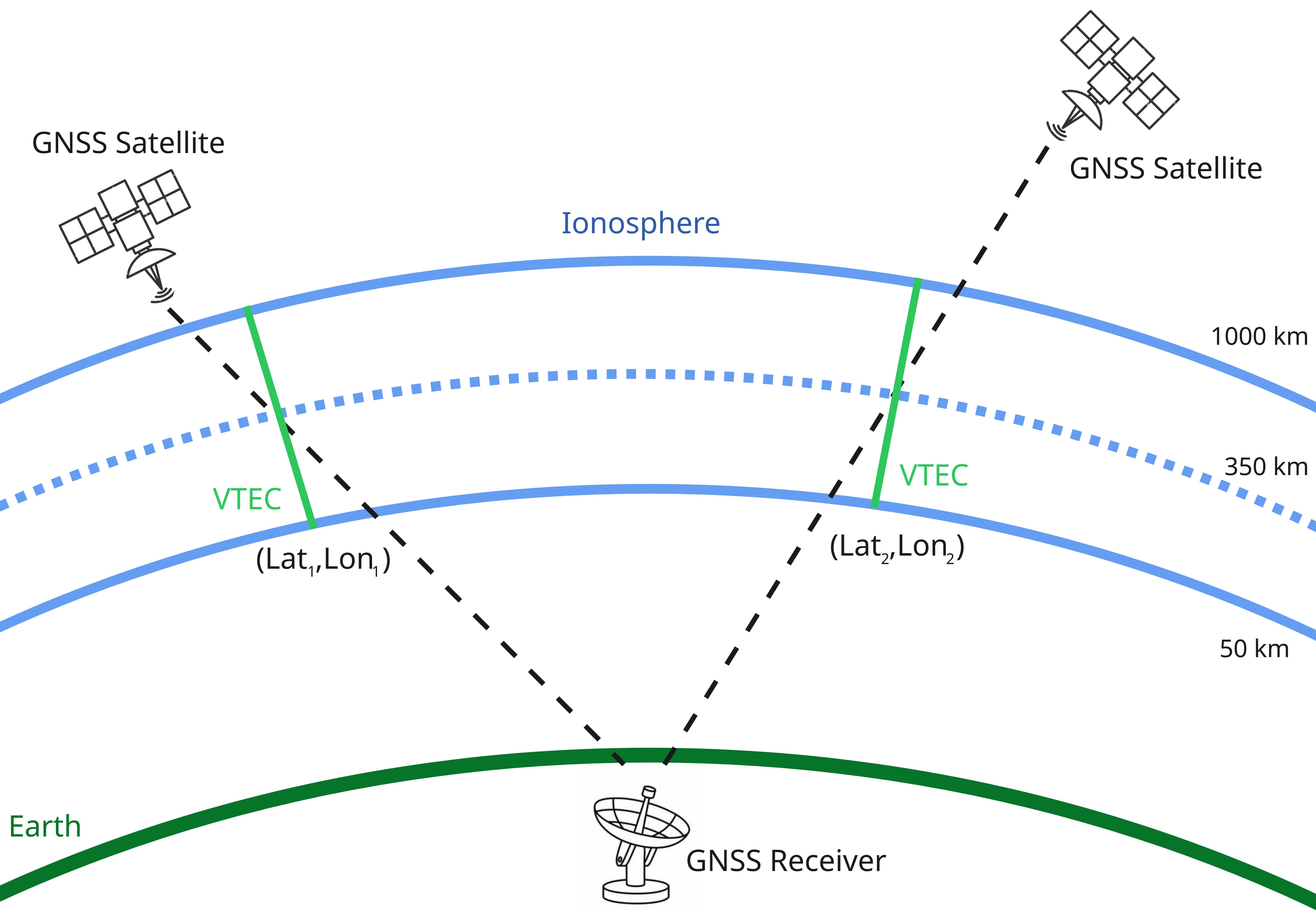}
    \caption{Schematic representation of the vTEC (green).}%Schematic cross-section showing the conversion of Total Electron Content (TEC) to Vertical Total Electron Content (vTEC, green) using the Single Layer Ionospheric Model. Two GPS satellites transmit signals through the ionospheric shell ($\sim$350 km altitude) to a ground receiver, with vTEC derived from TEC via the mapping function.
    \label{fig:TEC}
    \vspace{-10pt} % optional: remove vertical gap
\end{wrapfigure}
The ionosphere, Earth's charged atmospheric layer extending from $\sim$50 to 1,000 km altitude, plays a crucial role in modern technological infrastructure and daily life \cite{Berger2020}. This plasma layer, mainly influenced by solar radiation, magnetospheric convection and thermospheric neutral dynamics \cite{kelley2009}, directly affects critical systems including Global Navigation Satellite Systems (GNSS) accuracy, satellite communications, and Low Earth Orbit (LEO) orbital stability \cite{Kintner1976, Kataoka2022}. 
%Variations in electron density can cause significant range errors of up to 100 meters in GPS signals, disrupt satellite communications during geomagnetic storms, and affect everything from agricultural GPS guidance systems to shipping and transportation networks.
A fundamental parameter for characterizing ionospheric behavior is the Total Electron Content (TEC), defined as the integrated electron density along a given path. TEC data are often converted to vertical TEC (vTEC) using a shell approximation and mapping function to account for ray path obliquity \cite{Jakowski2011}. Figure \ref{fig:TEC} illustrates this process, showing slant paths from GNSS satellites projected to their vertical equivalent. Accurate vTEC prediction is essential for correcting GNSS signal delays, particularly during highly dynamic space weather events.

The Madrigal database \cite{Rideout}, maintained by CEDAR and hosted at MIT Haystack Observatory, is among the largest repositories of upper atmospheric data. It integrates multi-decade observations from over 159 instruments, including incoherent scatter radars and GNSS-based vTEC, providing global (albeit sparse) coverage with 5-minute temporal resolution. This unique dataset is used by the Jet Propulsion Laboratory (JPL) to generate Global Ionospheric Maps (GIM) of vTEC \cite{Martire2024} and provides an invaluable resource for machine learning applications in space weather research.

Despite rich observations, TEC prediction remains challenging due to nonlinear coupling between solar activity, geomagnetic conditions, and ionospheric response. Empirical models like the International Reference Ionosphere (IRI) \cite{Bilitza2011, Bilitza2017}, though useful for climatology, show reduced accuracy during geomagnetic disturbances when ionospheric dynamics are most critical. This motivates machine learning approaches that can capture complex patterns from multi-source data.

Here, we focus on accurate ionospheric TEC prediction by forecasting vTEC using a time-fusion transformer architecture, with the sparse GNSS Madrigal dataset serving as the target. Our framework provides a flexible benchmarking platform that supports probabilistic forecasting, heterogeneous input sources, and scientific experimentation by accommodating varying input features, time histories, and resolutions. To foster reproducibility and community-driven innovation, we publicly release our implementation as \texttt{ionopy}\footnote{\url{https://github.com/spaceml-org/ionopy}
, accessed 30 August 2025}, offering a foundation for both applied forecasting and fundamental ionospheric research.

%In this work, we address the critical need for accurate ionospheric TEC prediction by leveraging the sparse yet comprehensive Madrigal dataset, leveraging a temporal fusion transformer. Our approach enhances the reliability of satellite navigation and communication systems and offers a flexible platform for testing scientific hypotheses by incorporating additional data sources or varying input resolutions. The presented framework not only provides a way to train probabilistic machine learning transformer models from sparse data, but it also open-sources a framework~\footnote{\url{https://github.com/spaceml-org/ionopy}, date of access 30 August 2025.} where scientists and practictioners can run experiments leveraging different inputs, time histories, resolutions, etc. and test hypothesis.%The contribution represents a significant advancement in ionosphere prediction capabilities, with direct implications for improving the reliability and accuracy of satellite-based navigation and communication systems that billions of people depend on daily. Furthermore, our proposed framework can also be used to test scientific hypothesis by adding more data that can be considered more important, or by varying the time history or resolution of input data.

\section{Background}
\label{sec:background}

\subsection{Machine Learning and Ionosphere Modeling}
\label{sec:ML_and_ionosphere_modeling}

The application of machine learning to ionospheric modeling has gained significant momentum in recent years, driven by the limitations of traditional physics-based and empirical approaches and the increasing availability of datasets.  
Early studies have applied classical algorithms such as gradient boosting (XGBoost) and Multi-Layer Perceptrons (MLP) to predict the global state of the ionosphere \cite{Zhukov2021, Smirnov2023}, while more advanced methods, such as Bi-directional Long-Term-Short-Term Memory (BiLSTM) and transformers, have shown promising results in modeling temporal dependencies, though sometimes restricted to limited geographic areas. Part of these studies have leveraged the JPL-GIM as a target data source since it provides a homogeneous set with respect to latitude and longitude~\cite{shih2024forecasting}. Instead, some of the regional forecasting models have also used the sparse GNSS-derived vTEC and demonstrated that modern ML models trained on Madrigal TEC data could significantly outperform traditional models~\cite {Xiong2021, Yang2025}. 

The ability to ingest heterogeneous, multi-source data represents a key advantage of machine learning approaches over empirical and physics-based models. Modern ML models can potentially simultaneously incorporate solar activity indices (F10.7, sunspot numbers), geomagnetic activity parameters (Dst, Kp indices), solar wind parameters, historical TEC observations, or even SDO images, to create comprehensive predictive frameworks. This multi-modal capability is particularly important for ionospheric and thermosphere modeling, where the complex interplay between solar, magnetospheric, and thermospheric processes requires consideration of numerous input variables~\cite{acciarini2024improving}. 

\subsection{Temporal Fusion Transformers}
We employ Temporal Fusion Transformers (TFT)~\cite{lim2021temporal} to forecast vTEC behavior, leveraging their state-of-the-art design for multi-horizon time series forecasting with built-in interpretability. TFT integrates specialized components such as the variable selection network, sequence-to-sequence LSTM layers to replace positional encodings, and multi-head attention to handle heterogeneous inputs and temporal patterns common in real-world forecasting.

Key features include the variable selection network, which identifies and weights relevant input features, critical for ionospheric modeling, where parameter importance varies across conditions. Moreover, static covariate encoders incorporate time-invariant factors such as location, conditioning temporal processing with geographic and seasonal/daily contexts. Finally, the sequence-to-sequence LSTM layers replace positional encoding, better capturing local trends enriched with static information. Most importantly, TFT’s interpretable multi-head attention highlights influential time steps and features, offering transparent insights into relationships between solar activity, geomagnetic disturbances, and vTEC behavior. 
%TFT has shown superior performance over statistical and deep learning methods in several applications, handling missing data and multiple time scales effectively. These strengths make it well-suited for ionospheric TEC prediction using the sparse yet information-rich Madrigal dataset.

\section{Data \& Methodology}
\label{sec:methodology}

In terms of input features, day-of-year, longitude, and seconds-in-day are sine/cosine encoded to preserve cyclical relationships, while other coordinates are standardized or log-normalized depending on their statistical properties. Target vTEC values are log-transformed and standardized using pre-computed statistics to address typical skewness in ionospheric measurements.

The framework integrates diverse data sources with different temporal resolutions, including Thermosphere, Ionosphere, Mesosphere Energetics and Dynamics (TIMED) Solar EUV Experiment (SEE) Level 3, OMNI solar wind, indices and magnetic field data, JPLD Global Ionospheric Maps, solar activity proxies (F10.7, M10.7, S10.7, Y10.7), and geomagnetic indices (Dst, Ap). Temporal alignment and resampling are applied, followed by standardization or log-transformation for highly skewed data. JPLD data is subsampled to 10 equally spaced global points, reducing input features to 10. Static features (latitude, longitude, day-of-year) and temporal features (geomagnetic and solar inputs) are aligned, normalized, and prepared for input into a TFT model.

Input datasets include NASA's OMNIWeb 1-minute high-resolution data~\cite{king2005solar}, Ap index from Celestrack~\cite{bartels1949standardized, vallado2006using}, F10.7, M10.7, S10.7, Y10.7 solar proxies~\cite{tobiska2008development}, TIMED SEE Level 3 solar irradiance data up to 190nm at 1nm resolution~\cite{woods2005solar}, and JPL-GIM data~\cite{Martire2024}. For training (80\%), validation (10\%), and testing (10\%), a temporal split based on monthly rotations across years ensures models are evaluated on genuinely unseen periods while maintaining seasonal and solar activity representativeness, critical for assessing generalization to future space weather conditions. 

%This methodological framework enables the TFT architecture to effectively leverage the rich temporal and spatial structure present in the Madrigal dataset while maintaining scientific rigor in feature engineering and data handling practices essential for reliable ionospheric prediction models.

\section{Results}
\label{sec:results}

Table~\ref{tab:tft_results} summarizes the outcomes of experiments using the Temporal Fusion Transformer (TFT) with two attention heads, two LSTM layers, a hidden state size of 64, and a 10\% dropout rate. Each row of the table corresponds to a different experimental setup, where the input features, their historical time windows (lag), and temporal resolutions (res) were varied. Model performance was evaluated on a completely held-out test set spanning 2010–2025, using root mean squared error (RMSE) and mean absolute error (MAE). Both the mean ($\mu$) and standard deviation ($\sigma$) of vTEC (in TEC units, $1 \ \textrm{TECU}=10^{16}$ electrons/m$^2$) are reported, since these values are available in the Madrigal dataset and predicted by the TFT. Prediction horizons ranged from 60 minutes up to 24 hours.
\begin{figure}[tbh!]
    \centering
    \includegraphics[width=0.75\linewidth]{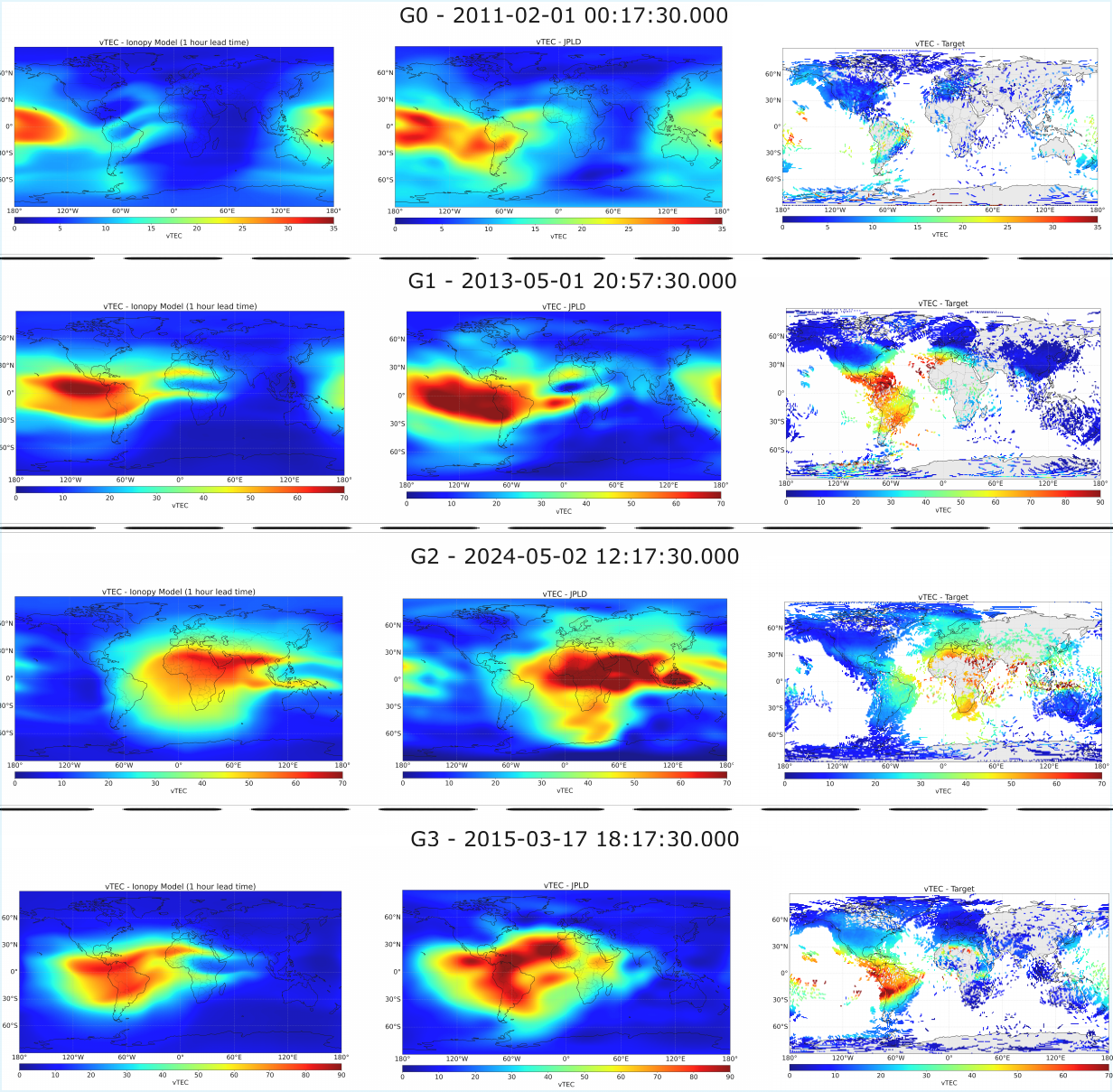}
    \caption{Results of Ionopy temporal fusion transformer model with 1 hour lead time (left column) against ground truth target vTEC data derived from GNSS (right column), and the corresponding calibrated JPL-GIM map (center column).}%Each row represents a different date in the unseen test set, with different geomagnetic storm scales, ranging from G0 (first row) to G3 (last row).}
    \label{fig:madrigal_test_set_inference_results}
\end{figure}%

\begin{table}[tbh!]
\caption{Temporal fusion transformer results with different input features lags and resolutions. Metrics include mean and standard deviation for MAE and RMSE on the unseen test set.}
\label{tab:tft_results}
\centering
\tiny
\begin{tabular}{ccccccccc}
\toprule
\textbf{\shortstack{OMNI \\ Indices \\ (min)}} & \textbf{\shortstack{OMNI \\ Solar \\ Wind\\ lag, res\\(min)}} & \textbf{\shortstack{OMNI \\ Magnetic \\ Field\\ lag, res\\(min)}} & \textbf{\shortstack{Ap \\ Index\\ lag, res\\(days)}} & \textbf{\shortstack{Solar \\ Proxies\\ lag, res\\(days)}} & \textbf{\shortstack{TIMED \\SEE \\ L3\\ lag, res\\(days)}} & \textbf{\shortstack{JPL-GIM\\ lag, res\\(min)}} & \textbf{\shortstack{MAE \\ $\mu$, $\sigma$ \\(TECU)}} & \textbf{\shortstack{RMSE \\ $\mu$, $\sigma$\\(TECU)}} \\ \midrule \midrule
1620, 60 & 1620, 60 & 1620, 60 & 27, 1 & \color{red}{\xmark} & 27, 1 & \color{red}{\xmark} & 2.37, 0.33 & 3.73, 0.51 \\
\color{red}{\xmark} & \color{red}{\xmark} & \color{red}{\xmark} & 81, 1 & \color{red}{\xmark} & 81, 1 & \color{red}{\xmark} & 2.47, 0.32 & 3.80, 0.50 \\
8640, 60 & 8640, 60 & 8640, 60 & 144, 1 & \color{red}{\xmark} & \color{red}{\xmark} & 8640, 60 & 2.16, 0.32 & 3.38, 0.49 \\
8640, 60 & \color{red}{\xmark} & 8640, 60 & 144, 1 & 144, 1 & \color{red}{\xmark} & 8640, 60 & 2.16, 0.31 & 3.40, 0.49 \\
8640, 60 & 8640, 60 & 8640, 60 & \color{red}{\xmark} & 144, 1 & \color{red}{\xmark} & 8640, 60 & 2.15, 0.31 & 3.37, 0.49 \\
8640, 60 & 8640, 60 & 8640, 60 & \color{red}{\xmark} & \color{red}{\xmark} & \color{red}{\xmark} & \color{red}{\xmark} & 3.85, 0.36 & 5.58, 0.55 \\
8640, 60 & 8640, 60 & 8640, 60 & 144, 1 & \color{red}{\xmark} & \color{red}{\xmark} & \color{red}{\xmark} & 3.47, 0.34 & 5.01, 0.51 \\
6480, 60 & 6480, 60 & 6480, 60 & 108, 1 & \color{red}{\xmark} & 108, 1 & \color{red}{\xmark} & 2.42, 0.33 & 3.81, 0.50 \\
3240, 60 & 3240, 60 & 3240, 60 & 54, 1 & \color{red}{\xmark} & 54, 1 & \color{red}{\xmark} & 2.38, 0.32 & 3.70, 0.50 \\
8100, 60 & 8100, 60 & 8100, 60 & 135, 1 & \color{red}{\xmark} & 135, 1 & \color{red}{\xmark} & 2.45, 0.32 & 3.86, 0.50 \\
8640, 60 & 8640, 60 & 8640, 60 & 144, 1 & 144, 1 & 144, 1 & \color{red}{\xmark} & 2.34, 0.32 & 3.67, 0.49 \\
4320, 60 & 4320, 60 & 4320, 60 & 72, 1 & \color{red}{\xmark} & 72, 1 & \color{red}{\xmark} & 2.41, 0.32 & 3.80, 0.50 \\
4860, 60 & 4860, 60 & 4860, 60 & 81, 1 & \color{red}{\xmark} & 81, 1 & \color{red}{\xmark} & 2.43, 0.32 & 3.79, 0.50 \\
8640, 60 & 8640, 60 & 8640, 60 & 144, 1 & 144, 1 & \color{red}{\xmark} & 8640, 60 & 2.17, 0.32 & 3.40, 0.50 \\
8640, 60 & 8640, 60 & 8640, 60 & 144, 1 & 144, 1 & \color{red}{\xmark}& \color{red}{\xmark} & 2.31, 0.31 & 3.60, 0.49 \\
8640, 60 & 8640, 60 & 8640, 60 & 144, 1 & \color{red}{\xmark} & 144, 1 & \color{red}{\xmark} & 2.35, 0.31 & 3.68, 0.49 \\
\color{red}{\xmark} & 8640, 60 & 8640, 60 & 144, 1 & 144, 1& 144, 1 & 8640, 60 & \textbf{2.13}, \textbf{0.31} & \textbf{3.33}, \textbf{0.49} \\
8640, 60 & 8640, 60 & \color{red}{\xmark} & 144, 1 & 144, 1 & 144, 1 & 8640, 60 & 2.14, 0.32 & 3.34, 0.49 \\
\bottomrule
\end{tabular}
\end{table}
Overall, models that incorporated the full set of features and longer historical time windows achieved the best performance, as expected. Importantly, the results demonstrate that TIMED SEE L3 irradiance data can effectively replace traditional solar proxies, providing comparable predictive accuracy while offering direct EUV irradiance measurements. Moreover, solar irradiance seems to be the most critical input for ionospheric forecasting: when JPL-GIM data, TIMED irradiance, and solar proxies were excluded, model errors nearly doubled (up to about 5.6 TECU in RMSE for the mean).
While the inclusion of JPL-GIM features provided a modest performance boost, the improvement was marginal. This suggests that EUV irradiance and geomagnetic activity alone are sufficient to enable reliable ionospheric forecasts up to one day ahead.

Figure~\ref{fig:madrigal_test_set_inference_results} further illustrates the TFT's predictions with a one-hour lead time (left column) compared against GNSS-derived ionospheric data (right column) under four different geomagnetic storm conditions (based on NOAA geomagnetic storm G-classes). The central column displays corresponding JPL-GIM maps (calibrated on sparse data, not predictions). Qualitatively, the model successfully reproduces key ionospheric structures, such as the equatorial double crest enhancement (also called Equatorial Ionization Anomaly) and day–night variations during geomagnetic storm conditions.

More detailed performance metrics are provided in Appendix, where Tables~\ref{tab:tft_latitude}, \ref{tab:tft_storm}, and \ref{tab:tft_solar} present loss values aggregated by latitude, geomagnetic storm conditions, and solar activity levels. %These results further highlight the role of different input features, lags, and resolutions under varying geographical and solar conditions.

\section{Conclusions}
\label{sec:conclusions}

In this work, we introduce and release the open-source framework \texttt{ionopy}, designed to preprocess, temporally align, and integrate solar and geomagnetic data to predict GNSS-derived sparse vTEC using Temporal Fusion Transformers. Solar EUV irradiance emerges as the most influential predictor, enabling accurate forecasts from 1 to 24 hours lead time, with RMSE as low as 3.33 TECU in the best experiments. These results highlight the potential of machine learning to outperform traditional ionospheric models and demonstrate the framework's flexibility for testing scientific hypotheses, including the impact of varying input features, lags, and temporal resolutions.
\section*{Acknowledgments}
This work is a research product of Heliolab (heliolab.ai), an initiative of the Frontier Development Lab (FDL.ai), delivered by Trillium Technologies in partnership with NASA, Google Cloud, and NVIDIA. The authors also thank NASA's Jet Propulsion Laboratory for their continuing support. 
 
%%%%%%%%%%%%%%%%%%%%%%%%%%%%%%%%%%%%%%%%%%%%%%%%%%%%%%%%%%%%
\clearpage
\newpage
\appendix

%%%%%%%%%%%%%%%%%%%%%%%%%%%%%%%%%%%%%%%%%%%%%%%%%%%%%%%%%%%%

\pagebreak
\bibliographystyle{unsrt}
\bibliography{reference.bib}

\section*{Appendix}
\begin{table}[tbh!]
\caption{TFT results for different latitude bands. Metrics include MAE and RMSE for both mean and standard deviation.}
\label{tab:tft_latitude}
\centering
\tiny
\begin{tabular}{cccccccc}
\toprule
\textbf{\shortstack{OMNI \\ Indices \\ (min)}} & \textbf{\shortstack{OMNI \\ Solar \\ Wind\\ lag, res\\(min)}} & \textbf{\shortstack{OMNI \\ Magnetic \\ Field\\ lag, res\\(min)}} & \textbf{\shortstack{Ap \\ Index\\ lag, res\\(days)}} & \textbf{\shortstack{Solar \\ Proxies\\ lag, res\\(days)}} & \textbf{\shortstack{TIMED \\SEE \\ L3\\ lag, res\\(days)}} & \textbf{\shortstack{JPL-GIM\\ lag, res\\(min)}} & \textbf{\shortstack{RMSE \\ $\mu$, $\sigma$\\(TECU)}} \\ \midrule \midrule
1620, 60 & 1620, 60 & 1620, 60 & 27, 1 & \color{red}{\xmark} & 27, 1 & \color{red}{\xmark} & \begin{tabular}{c}
<30°: 5.663, 0.578 \\
30–60°: 2.625, 0.407 \\
>60°: 2.217, 0.511
\end{tabular}  \\
\hline
\color{red}{\xmark} & \color{red}{\xmark} & \color{red}{\xmark} & 81, 1 & \color{red}{\xmark} & 81, 1 & \color{red}{\xmark} & \begin{tabular}{c}
<30°: 5.707, 0.570 \\
30–60°: 2.719, 0.396 \\
>60°: 2.338, 0.501
\end{tabular} \\
\hline
8640, 60 & 8640, 60 & 8640, 60 & 144, 1 & \color{red}{\xmark} & \color{red}{\xmark} & 8640, 60 & \begin{tabular}{c}
<30°: 5.135, 0.568 \\
30–60°: 2.341, 0.395 \\
>60°: 2.100, 0.498
\end{tabular} \\
\hline
8640, 60 & \color{red}{\xmark} & 8640, 60 & 144, 1 & 144, 1 & \color{red}{\xmark} & 8640, 60 & \begin{tabular}{c}
<30°: 5.159, 0.563 \\
30–60°: 2.365, 0.394 \\
>60°: \textbf{2.077}, 0.496
\end{tabular} \\
\hline
8640, 60 & 8640, 60 & 8640, 60 & \color{red}{\xmark} & 144, 1 & \color{red}{\xmark} & 8640, 60 & \begin{tabular}{c}
<30°: 5.084, 0.563 \\
30–60°: 2.353, 0.393 \\
>60°: 2.107, 0.494
\end{tabular} \\
\hline
8640, 60 & 8640, 60 & 8640, 60 & \color{red}{\xmark} & \color{red}{\xmark} & \color{red}{\xmark} & \color{red}{\xmark} & \begin{tabular}{c}
<30°: 8.220, 0.620 \\
30–60°: 4.358, 0.452 \\
>60°: 3.216, 0.553
\end{tabular} \\
\hline
8640, 60 & 8640, 60 & 8640, 60 & 144, 1 & \color{red}{\xmark} & \color{red}{\xmark} & \color{red}{\xmark} & \begin{tabular}{c}
<30°: 7.370, 0.590 \\
30–60°: 3.900, 0.416 \\
>60°: 2.945, 0.519
\end{tabular} \\
\hline
6480, 60 & 6480, 60 & 6480, 60 & 108, 1 & \color{red}{\xmark} & 108, 1 & \color{red}{\xmark} &\begin{tabular}{c}
<30°: 5.826, 0.572 \\
30–60°: 2.644, 0.405 \\
>60°: 2.285, 0.497
\end{tabular} \\
\hline
3240, 60 & 3240, 60 & 3240, 60 & 54, 1 & \color{red}{\xmark} & 54, 1 & \color{red}{\xmark} & \begin{tabular}{c}
<30°: 5.591, 0.571 \\
30–60°: 2.643, 0.400 \\
>60°: 2.249, 0.501
\end{tabular} \\
\hline
8100, 60 & 8100, 60 & 8100, 60 & 135, 1 & \color{red}{\xmark} & 135, 1 & \color{red}{\xmark} & \begin{tabular}{c}
<30°: 5.893, 0.572 \\
30–60°: 2.692, 0.399 \\
>60°: 2.271, 0.500
\end{tabular} \\
\hline
8640, 60 & 8640, 60 & 8640, 60 & 144, 1 & 144, 1 & 144, 1 & \color{red}{\xmark} & \begin{tabular}{c}
<30°: 5.562, 0.569 \\
30–60°: 2.579, 0.397 \\
>60°: 2.229, 0.496
\end{tabular}  \\
\hline
4320, 60 & 4320, 60 & 4320, 60 & 72, 1 & \color{red}{\xmark} & 72, 1 & \color{red}{\xmark} & \begin{tabular}{c}
<30°: 5.801, 0.572 \\
30–60°: 2.640, 0.400 \\
>60°: 2.269, 0.497
\end{tabular} \\
\hline
4860, 60 & 4860, 60 & 4860, 60 & 81, 1 & \color{red}{\xmark} & 81, 1 & \color{red}{\xmark} & \begin{tabular}{c}
<30°: 5.756, 0.572 \\
30–60°: 2.689, 0.403 \\
>60°: 2.256, 0.501
\end{tabular}  \\
\hline
8640, 60 & 8640, 60 & 8640, 60 & 144, 1 & 144, 1 & \color{red}{\xmark} & 8640, 60 & \begin{tabular}{c}
<30°: 5.155, 0.574 \\
30–60°: 2.371, 0.401 \\
>60°: 2.106, 0.498
\end{tabular}  \\
\hline
8640, 60 & 8640, 60 & 8640, 60 & 144, 1 & 144, 1 & \color{red}{\xmark}& \color{red}{\xmark} & \begin{tabular}{c}
<30°: 5.450, \textbf{0.559} \\
30–60°: 2.555, \textbf{0.389} \\
>60°: 2.179, \textbf{0.491}
\end{tabular}  \\
\hline
8640, 60 & 8640, 60 & 8640, 60 & 144, 1 & \color{red}{\xmark} & 144, 1 & \color{red}{\xmark} & \begin{tabular}{c}
<30°: 5.569, 0.567 \\
30–60°: 2.594, 0.395 \\
>60°: 2.219, 0.494
\end{tabular} \\
\hline
\color{red}{\xmark} & 8640, 60 & 8640, 60 & 144, 1 & 144, 1& 144, 1 & 8640, 60 & \begin{tabular}{c}
<30°: 5.041, 0.561 \\
30–60°: \textbf{2.319}, 0.393 \\
>60°: 2.093, 0.493
\end{tabular} \\
\hline
8640, 60 & 8640, 60 & \color{red}{\xmark} & 144, 1 & 144, 1 & 144, 1 & 8640, 60 & \begin{tabular}{c}
<30°: \textbf{5.037}, 0.560 \\
30–60°: 2.335, 0.390 \\
>60°: 2.100, 0.493
\end{tabular} \\
\bottomrule
\end{tabular}
\end{table}

\begin{table}[tbh!]
\caption{TFT results for different geomagnetic storm conditions. Metrics include MAE and RMSE for both mean and standard deviation.
}
\label{tab:tft_storm}
\centering
\tiny
\begin{tabular}{cccccccc}
\toprule
\textbf{\shortstack{OMNI \\ Indices \\ (min)}} & \textbf{\shortstack{OMNI \\ Solar \\ Wind\\ lag, res\\(min)}} & \textbf{\shortstack{OMNI \\ Magnetic \\ Field\\ lag, res\\(min)}} & \textbf{\shortstack{Ap \\ Index\\ lag, res\\(days)}} & \textbf{\shortstack{Solar \\ Proxies\\ lag, res\\(days)}} & \textbf{\shortstack{TIMED \\SEE \\ L3\\ lag, res\\(days)}} & \textbf{\shortstack{JPL-GIM\\ lag, res\\(min)}} & \textbf{\shortstack{RMSE \\ $\mu$, $\sigma$\\(TECU)}} \\ \midrule \midrule
1620, 60 & 1620, 60 & 1620, 60 & 27, 1 & \color{red}{\xmark} & 27, 1 & \color{red}{\xmark} & 
\begin{tabular}{l}
Ap [0-39]: 3.64, 0.50 \\
Ap (39-67]: 6.86, 0.76 \\
Ap (67-111]: 8.55, 1.08 \\
Ap (111-300]: 10.43, 0.74 \\
\end{tabular}
\\
\hline
\color{red}{\xmark} & \color{red}{\xmark} & \color{red}{\xmark} & 81, 1 & \color{red}{\xmark} & 81, 1 & \color{red}{\xmark} & 
\begin{tabular}{l}
Ap [0-39]: 3.70, 0.49 \\
Ap (39-67]: 7.94, 0.77 \\
Ap (67-111]: 8.98, 1.13 \\
Ap (111-300]: 11.01, 0.74 \\
\end{tabular}
\\
\hline
8640, 60 & 8640, 60 & 8640, 60 & 144, 1 & \color{red}{\xmark} & \color{red}{\xmark} & 8640, 60 &
\begin{tabular}{l}
Ap [0-39]: 3.31, 0.49 \\
Ap (39-67]: 6.66, 0.76 \\
Ap (67-111]: 7.27, 1.08 \\
Ap (111-300]: \textbf{9.28}, 0.74 \\
\end{tabular}
 \\
\hline
8640, 60 & \color{red}{\xmark} & 8640, 60 & 144, 1 & 144, 1 & \color{red}{\xmark} & 8640, 60 & 
\begin{tabular}{l}
Ap [0-39]: 3.32, 0.48 \\
Ap (39-67]: 6.72, 0.77 \\
Ap (67-111]: \textbf{7.14}, 1.12 \\
Ap (111-300]: 9.32, 0.73 \\
\end{tabular}

\\
\hline
8640, 60 & 8640, 60 & 8640, 60 & \color{red}{\xmark} & 144, 1 & \color{red}{\xmark} & 8640, 60 & 
\begin{tabular}{l}
Ap [0-39]: 3.29, 0.48 \\
Ap (39-67]: \textbf{6.46}, 0.76 \\
Ap (67-111]: 7.30, 1.13 \\
Ap (111-300]: 9.34, \textbf{0.73} \\
\end{tabular}
 \\
\hline
8640, 60 & 8640, 60 & 8640, 60 & \color{red}{\xmark} & \color{red}{\xmark} & \color{red}{\xmark} & \color{red}{\xmark} & 
\begin{tabular}{l}
Ap [0-39]: 5.49, 0.54 \\
Ap (39-67]: 8.74, 0.84 \\
Ap (67-111]: 9.92, 1.19 \\
Ap (111-300]: 12.33, 0.78 \\
\end{tabular}
\\
\hline
8640, 60 & 8640, 60 & 8640, 60 & 144, 1 & \color{red}{\xmark} & \color{red}{\xmark} & \color{red}{\xmark} &
\begin{tabular}{l}
Ap [0-39]: 4.94, 0.51 \\
Ap (39-67]: 7.68, 0.75 \\
Ap (67-111]: 8.35, 1.08 \\
Ap (111-300]: 10.95, 0.74 \\
\end{tabular}
\\
\hline
6480, 60 & 6480, 60 & 6480, 60 & 108, 1 & \color{red}{\xmark} & 108, 1 & \color{red}{\xmark} &
\begin{tabular}{l}
Ap [0-39]: 3.73, 0.49 \\
Ap (39-67]: 7.41, 0.76 \\
Ap (67-111]: 8.36, 1.11 \\
Ap (111-300]: 10.45, 0.75 \\
\end{tabular}
\\
\hline
3240, 60 & 3240, 60 & 3240, 60 & 54, 1 & \color{red}{\xmark} & 54, 1 & \color{red}{\xmark} & 
\begin{tabular}{l}
Ap [0-39]: 3.62, 0.49 \\
Ap (39-67]: 7.17, 0.76 \\
Ap (67-111]: 8.55, 1.12 \\
Ap (111-300]: 9.91, 0.74 \\
\end{tabular}
\\
\hline
8100, 60 & 8100, 60 & 8100, 60 & 135, 1 & \color{red}{\xmark} & 135, 1 & \color{red}{\xmark} &
\begin{tabular}{l}
Ap [0-39]: 3.77, 0.49 \\
Ap (39-67]: 7.25, 0.77 \\
Ap (67-111]: 8.62, 1.13 \\
Ap (111-300]: 10.82, 0.74 \\
\end{tabular}
\\
\hline
8640, 60 & 8640, 60 & 8640, 60 & 144, 1 & 144, 1 & 144, 1 & \color{red}{\xmark} & 
\begin{tabular}{l}
Ap [0-39]: 3.59, 0.49 \\
Ap (39-67]: 6.93, 0.76 \\
Ap (67-111]: 8.33, 1.12 \\
Ap (111-300]: 9.90, 0.74 \\
\end{tabular}
\\
\hline
4320, 60 & 4320, 60 & 4320, 60 & 72, 1 & \color{red}{\xmark} & 72, 1 & \color{red}{\xmark} & 
\begin{tabular}{l}
Ap [0-39]: 3.71, 0.49 \\
Ap (39-67]: 7.47, 0.77 \\
Ap (67-111]: 8.65, 1.12 \\
Ap (111-300]: 9.84, 0.74 \\
\end{tabular}
\\
\hline
4860, 60 & 4860, 60 & 4860, 60 & 81, 1 & \color{red}{\xmark} & 81, 1 & \color{red}{\xmark} & 
\begin{tabular}{l}
Ap [0-39]: 3.70, 0.49 \\
Ap (39-67]: 7.21, 0.77 \\
Ap (67-111]: 8.57, 1.12 \\
Ap (111-300]: 9.97, 0.75 \\
\end{tabular}
 \\
\hline
8640, 60 & 8640, 60 & 8640, 60 & 144, 1 & 144, 1 & \color{red}{\xmark} & 8640, 60 & 
\begin{tabular}{l}
Ap [0-39]: 3.32, 0.49 \\
Ap (39-67]: 6.85, 0.74 \\
Ap (67-111]: 7.38, 1.12 \\
Ap (111-300]: 9.44, 0.74 \\
\end{tabular}
 \\
\hline
8640, 60 & 8640, 60 & 8640, 60 & 144, 1 & 144, 1 & \color{red}{\xmark}& \color{red}{\xmark} &
\begin{tabular}{l}
Ap [0-39]: 3.52, 0.48 \\
Ap (39-67]: 6.86, 0.77 \\
Ap (67-111]: 8.42, 1.10 \\
Ap (111-300]: 10.12, 0.73 \\
\end{tabular}
\\
\hline
8640, 60 & 8640, 60 & 8640, 60 & 144, 1 & \color{red}{\xmark} & 144, 1 & \color{red}{\xmark} & 
\begin{tabular}{l}
Ap [0-39]: 3.59, 0.49 \\
Ap (39-67]: 7.03, 0.75 \\
Ap (67-111]: 8.39, 1.09 \\
Ap (111-300]: 10.10, 0.74 \\
\end{tabular}
 \\
\hline
\color{red}{\xmark} & 8640, 60 & 8640, 60 & 144, 1 & 144, 1& 144, 1 & 8640, 60 & 
\begin{tabular}{l}
Ap [0-39]: \textbf{3.25}, \textbf{0.48} \\
Ap (39-67]: 6.83, 0.78 \\
Ap (67-111]: 7.24, 1.14 \\
Ap (111-300]: 9.52, 0.74 \\
\end{tabular}
 \\
\hline
8640, 60 & 8640, 60 & \color{red}{\xmark} & 144, 1 & 144, 1 & 144, 1 & 8640, 60 & 
\begin{tabular}{l}
Ap [0-39]: 3.26, 0.48 \\
Ap (39-67]: 6.64, \textbf{0.74} \\
Ap (67-111]: 7.31, \textbf{1.08} \\
Ap (111-300]: 9.54, 0.74 \\
\end{tabular}
 \\
\bottomrule
\end{tabular}
\end{table}

\begin{table}[tbh!]
\caption{TFT results for different solar irradiance activity levels. Metrics include MAE and RMSE for both mean and standard deviation.}
\label{tab:tft_solar}
\centering
\tiny
\begin{tabular}{cccccccc}
\toprule
\textbf{\shortstack{OMNI \\ Indices \\ (min)}} & \textbf{\shortstack{OMNI \\ Solar \\ Wind\\ lag, res\\(min)}} & \textbf{\shortstack{OMNI \\ Magnetic \\ Field\\ lag, res\\(min)}} & \textbf{\shortstack{Ap \\ Index\\ lag, res\\(days)}} & \textbf{\shortstack{Solar \\ Proxies\\ lag, res\\(days)}} & \textbf{\shortstack{TIMED \\SEE \\ L3\\ lag, res\\(days)}} & \textbf{\shortstack{JPL-GIM\\ lag, res\\(min)}} & \textbf{\shortstack{RMSE \\ $\mu$, $\sigma$\\(TECU)}} \\ \midrule \midrule
1620, 60 & 1620, 60 & 1620, 60 & 27, 1 & \color{red}{\xmark} & 27, 1 & \color{red}{\xmark} & 
\begin{tabular}{l}
F10.7 [0-70]: 1.88, 0.28 \\
F10.7 (70-150]: 3.62, 0.53 \\
F10.7 [150-200): 5.73, 0.66 \\
\end{tabular}
\\
\hline
\color{red}{\xmark} & \color{red}{\xmark} & \color{red}{\xmark} & 81, 1 & \color{red}{\xmark} & 81, 1 & \color{red}{\xmark} & 
\begin{tabular}{l}
F10.7 [0-70]: 1.88, 0.28 \\
F10.7 (70-150]: 3.68, 0.52 \\
F10.7 [150-200): 5.87, 0.65 \\
\end{tabular}
\\
\hline
8640, 60 & 8640, 60 & 8640, 60 & 144, 1 & \color{red}{\xmark} & \color{red}{\xmark} & 8640, 60 &
\begin{tabular}{l}
F10.7 [0-70]: 1.72, 0.27 \\
F10.7 (70-150]: 3.28, 0.51 \\
F10.7 [150-200): 5.20, 0.67 \\
\end{tabular}
 \\
\hline
8640, 60 & \color{red}{\xmark} & 8640, 60 & 144, 1 & 144, 1 & \color{red}{\xmark} & 8640, 60 & 
\begin{tabular}{l}
F10.7 [0-70]: 1.72, 0.27 \\
F10.7 (70-150]: 3.28, 0.51 \\
F10.7 [150-200): 5.24, 0.66 \\
\end{tabular}
\\
\hline
8640, 60 & 8640, 60 & 8640, 60 & \color{red}{\xmark} & 144, 1 & \color{red}{\xmark} & 8640, 60 & 
\begin{tabular}{l}
F10.7 [0-70]: 1.73, 0.27 \\
F10.7 (70-150]: 3.26, 0.51 \\
F10.7 [150-200): 5.14, 0.65 \\
\end{tabular}
 \\
\hline
8640, 60 & 8640, 60 & 8640, 60 & \color{red}{\xmark} & \color{red}{\xmark} & \color{red}{\xmark} & \color{red}{\xmark} & 
\begin{tabular}{l}
F10.7 [0-70]: 2.33, 0.31 \\
F10.7 (70-150]: 5.42, 0.58 \\
F10.7 [150-200): 8.98, 0.69 \\
\end{tabular}
\\
\hline
8640, 60 & 8640, 60 & 8640, 60 & 144, 1 & \color{red}{\xmark} & \color{red}{\xmark} & \color{red}{\xmark} &
\begin{tabular}{l}
F10.7 [0-70]: 2.12, 0.31 \\
F10.7 (70-150]: 4.95, 0.53 \\
F10.7 [150-200): 8.27, 0.67 \\
\end{tabular}
\\
\hline
6480, 60 & 6480, 60 & 6480, 60 & 108, 1 & \color{red}{\xmark} & 108, 1 & \color{red}{\xmark} &
\begin{tabular}{l}
F10.7 [0-70]: 1.86, 0.28 \\
F10.7 (70-150]: 3.70, 0.52 \\
F10.7 [150-200): 5.91, 0.66 \\
\end{tabular}
\\
\hline
3240, 60 & 3240, 60 & 3240, 60 & 54, 1 & \color{red}{\xmark} & 54, 1 & \color{red}{\xmark} & 
\begin{tabular}{l}
F10.7 [0-70]: 1.84, 0.28 \\
F10.7 (70-150]: 3.59, 0.52 \\
F10.7 [150-200): 5.79, 0.66 \\
\end{tabular}
\\
\hline
8100, 60 & 8100, 60 & 8100, 60 & 135, 1 & \color{red}{\xmark} & 135, 1 & \color{red}{\xmark} &
\begin{tabular}{l}
F10.7 [0-70]: 1.87, 0.28 \\
F10.7 (70-150]: 3.73, 0.52 \\
F10.7 [150-200): 6.04, 0.65 \\
\end{tabular}
\\
\hline
8640, 60 & 8640, 60 & 8640, 60 & 144, 1 & 144, 1 & 144, 1 & \color{red}{\xmark} & 
\begin{tabular}{l}
F10.7 [0-70]: 1.86, 0.28 \\
F10.7 (70-150]: 3.58, 0.52 \\
F10.7 [150-200): 5.60, 0.65 \\
\end{tabular}
\\
\hline
4320, 60 & 4320, 60 & 4320, 60 & 72, 1 & \color{red}{\xmark} & 72, 1 & \color{red}{\xmark} & 
\begin{tabular}{l}
F10.7 [0-70]: 1.86, 0.28 \\
F10.7 (70-150]: 3.70, 0.52 \\
F10.7 [150-200): 5.87, 0.66 \\
\end{tabular}
\\
\hline
4860, 60 & 4860, 60 & 4860, 60 & 81, 1 & \color{red}{\xmark} & 81, 1 & \color{red}{\xmark} & 
\begin{tabular}{l}
F10.7 [0-70]: 1.88, 0.28 \\
F10.7 (70-150]: 3.65, 0.52 \\
F10.7 [150-200): 5.97, 0.66 \\
\end{tabular}
 \\
\hline
8640, 60 & 8640, 60 & 8640, 60 & 144, 1 & 144, 1 & \color{red}{\xmark} & 8640, 60 & 
\begin{tabular}{l}
F10.7 [0-70]: 1.72, 0.28 \\
F10.7 (70-150]: 3.29, 0.52 \\
F10.7 [150-200): 5.23, 0.68 \\
\end{tabular}
 \\
\hline
8640, 60 & 8640, 60 & 8640, 60 & 144, 1 & 144, 1 & \color{red}{\xmark}& \color{red}{\xmark} &
\begin{tabular}{l}
F10.7 [0-70]: 1.81, 0.27 \\
F10.7 (70-150]: 3.49, 0.51 \\
F10.7 [150-200): 5.56, 0.65 \\
\end{tabular}
\\
\hline
8640, 60 & 8640, 60 & 8640, 60 & 144, 1 & \color{red}{\xmark} & 144, 1 & \color{red}{\xmark} & 
\begin{tabular}{l}
F10.7 [0-70]: 1.84, 0.28 \\
F10.7 (70-150]: 3.57, 0.51 \\
F10.7 [150-200): 5.70, 0.65 \\
\end{tabular}
 \\
\hline
\color{red}{\xmark} & 8640, 60 & 8640, 60 & 144, 1 & 144, 1& 144, 1 & 8640, 60 & 
\begin{tabular}{l}
F10.7 [0-70]: \textbf{1.69}, \textbf{0.27} \\
F10.7 (70-150]: \textbf{3.22}, \textbf{0.51} \\
F10.7 [150-200): 5.14, 0.65 \\
\end{tabular}
 \\
\hline
8640, 60 & 8640, 60 & \color{red}{\xmark} & 144, 1 & 144, 1 & 144, 1 & 8640, 60 & 
\begin{tabular}{l}
F10.7 [0-70]: 1.72, 0.27 \\
F10.7 (70-150]: 3.23, 0.51 \\
F10.7 [150-200): \textbf{5.12}, \textbf{0.64} \\
\end{tabular}
 \\
\bottomrule
\end{tabular}
\end{table}

\end{document}